\documentclass{article}

\usepackage{arxiv}

\usepackage[utf8]{inputenc} 
\usepackage[T1]{fontenc}    
\usepackage{hyperref}       
\usepackage{url}            
\usepackage{booktabs}       
\usepackage{amsfonts}       
\usepackage{nicefrac}       
\usepackage{microtype}      
\usepackage{lipsum}
\usepackage{graphicx}
\graphicspath{ {./images/} }

\usepackage{amsmath, amsfonts}
\usepackage{algorithmic}
\usepackage{algorithm}
\usepackage{cite}
\usepackage{array}
\usepackage[caption=false, font=footnotesize]{subfig}
\usepackage{textcomp}
\usepackage{dblfloatfix}
\usepackage{url}
\usepackage{graphicx}
\usepackage{xcolor}
\usepackage{tikz}
\usepackage{pgfplots}
\usepackage{xcolor}
\usepackage{adjustbox}
\usepackage{booktabs}
\usepackage{multirow}
\usepackage{caption}
\pgfplotsset{compat=1.18}
\usetikzlibrary{positioning, decorations.pathreplacing}

\title{Few-Shot Inspired Generative Zero-Shot Learning}

\author{
 Md Shakil Ahamed Shohag \\
  Department of Electrical and Computer Engineering\\
  University of Windsor\\
  Canada \\
  \texttt{shakilshohag@gmail.com} \\
   \And
 Q. M. Jonathan Wu \\
  Department of Electrical and Computer Engineering\\
  University of Windsor\\
  Canada \\
  \texttt{jwu@uwindsor.ca} \\
  \And
 Farhad Pourpanah \\
  Department of Electrical and Computer Engineering\\
  Queen’s University\\
  Canada \\
  \texttt{farhad.086@gmail.com} \\
}

\begin{document}
\maketitle
\begin{abstract}
Generative zero-shot learning (ZSL) methods typically synthesize visual features for unseen classes using predefined semantic attributes, followed by training a fully supervised classification model. While effective, these methods require substantial computational resources and extensive synthetic data, thereby relaxing the original ZSL assumptions.
In this paper, we propose FSIGenZ, a few-shot-inspired generative ZSL framework that reduces reliance on large-scale feature synthesis. Our key insight is that class-level attributes exhibit instance-level variability, i.e., some attributes may be absent or partially visible, yet conventional ZSL methods treat them as uniformly present. To address this, we introduce Model-Specific Attribute Scoring (MSAS), which dynamically re-scores class attributes based on model-specific optimization to approximate instance-level variability without access to unseen data. We further estimate group-level prototypes as clusters of instances based on MSAS-adjusted attribute scores, which serve as representative synthetic features for each unseen class. To mitigate the resulting data imbalance, we introduce a Dual-Purpose Semantic Regularization (DPSR) strategy while training a semantic-aware contrastive classifier (SCC) using these prototypes. Experiments on SUN, AwA2, and CUB benchmarks demonstrate that FSIGenZ achieves competitive performance using far fewer synthetic features. 
\end{abstract}

\keywords{Zero-shot learning \and generalized zero-shot learning \and knowledge transfer \and contrastive learning \and feature synthesis}

\section{Introduction}
\label{sec:intro}
Zero-shot learning (ZSL) aims to recognize unseen classes by transferring knowledge from seen classes using semantic information, such as attributes or word vectors \cite{xian2018zero, lampert2013attribute}. This is crucial for tasks where collecting labeled data for every class is impractical. ZSL methods can be categorized by classification range: \textit{(i)} conventional ZSL (CZSL), which predicts only unseen classes, and \textit{(ii)} generalized ZSL (GZSL), which considers both seen and unseen classes during inference \cite{pourpanah2022review}.

\begin{figure}[t]
\centering
  \includegraphics[width=0.5\textwidth]{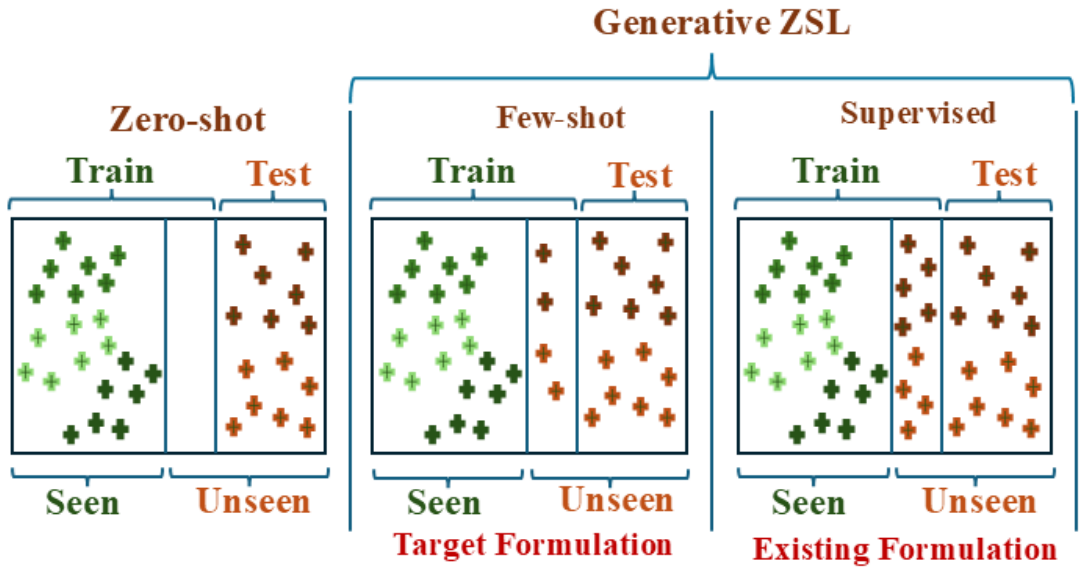}
  \caption {Conceptual illustration: FSL is a less relaxed problem formulation of ZSL than the supervised formulation.}
  \label{fig:intro}
\end{figure}

Early ZSL methods focus on learning a shared embedding space that maps visual features of seen classes to their corresponding semantic representations, enabling classification via similarity-based matching \cite{lampert2013attribute}. Although effective in transferring knowledge, these embedding-based methods often exhibit a strong bias toward seen classes in the GZSL setting, which leads to misclassifications of unseen instances \cite{chen2023duet}. To mitigate this bias, generative approaches leverage models such as generative adversarial networks (GANs) \cite{goodfellow2020generative} and variational autoencoders (VAEs) \cite{kingma2013auto} to synthesize visual features for unseen classes using semantic and visual information from seen classes. This strategy transforms ZSL into a fully supervised learning task and reduces the seen-class bias \cite{xian2018feature}. However, existing generative methods often synthesize a large volume of features without assessing their actual utility for classification \cite{gowda2023synthetic}. Moreover, this large-scale feature generation represents a highly relaxed ZSL setting and overlooks more constrained and realistic scenarios, such as few-shot learning (FSL), where only a limited number of examples per unseen class are available (see Fig.~\ref{fig:intro}).


\begin{figure}[t]
\centering
  \includegraphics[width=0.50\textwidth]{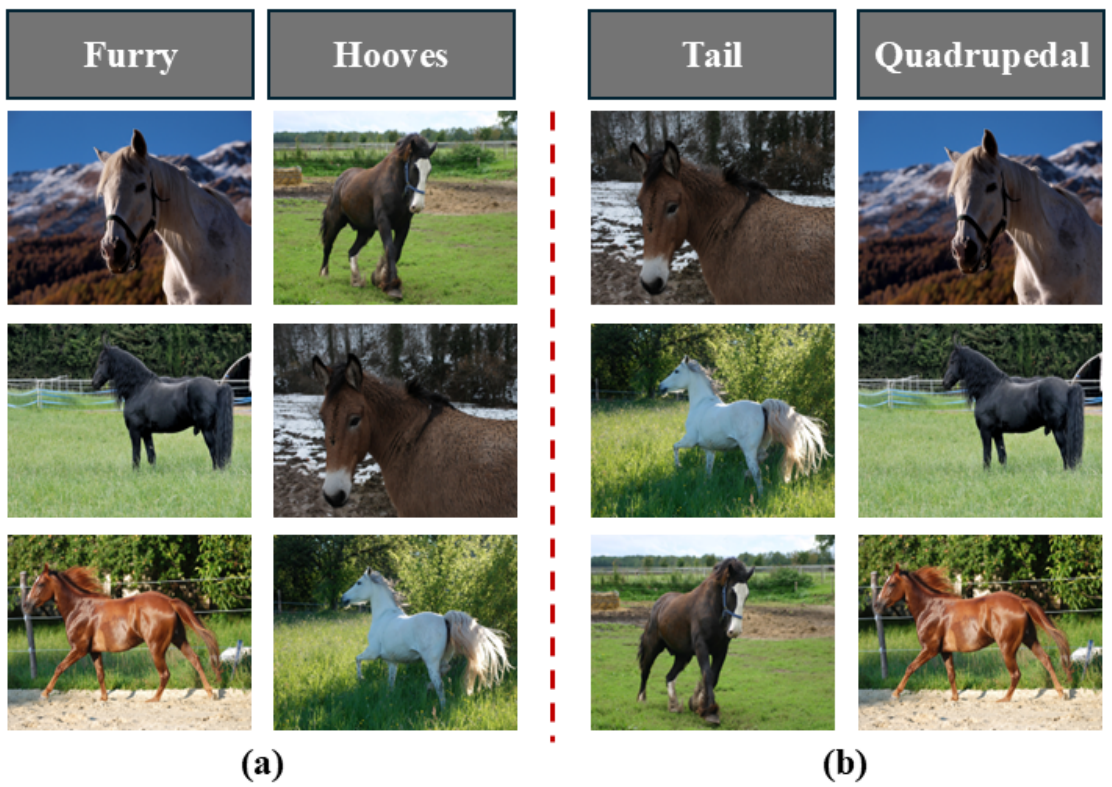}
  \caption {Our observations: Even for four attributes—Furry, Hooves, Tail, and Quadrupedal—(a) shows ungrouped instances of the horses as if they share identical attribute scores, while (b) conceptualizes how instances can be grouped (each row) based on varying degrees of attribute presence or absence.}
  \label{fig:msas}
\end{figure}

In our analysis, we found that many images exhibit missing or partially present attributes (see Fig. \ref{fig:msas}(a)). For example, while 'Furry' is consistently present, attributes such as 'Quadrupedal,' 'Tail,' and 'Hooves' are absent or only partially visible. This instance-level variability is overlooked by class-level semantics. Based on this observation, we propose two key insights: (\textit{i}) attribute scores should be dynamic to reflect the degree of presence in individual instances, and (\textit{ii}) instances can be meaningfully grouped based on these attribute scores (see Fig.\ref{fig:msas}(b)). However, implementing instance-level attribute scoring is challenging in ZSL as unseen classes lack instance-level data during training, making it impossible to capture attribute variations directly.

To overcome this, we propose \textit{Model-Specific Attribute Scoring (MSAS)} (Section \ref{secsec:msas}), which re-scores class-level attributes to model-optimized values by treating them as hyperparameters. This enables the model to better align with the true attribute variability without requiring access to unseen class instances. Building on MSAS, we estimate centroids of these attribute-based groups (Fig.~\ref{fig:msas}(b)) to serve as prototypes for each unseen class. These group-level prototypes act as representative instances during training and eliminate the need for large-scale feature generation that helps reduce computational overhead. To further address the class imbalance caused by this reduced feature synthesis, we employ a semantic-aware contrastive classifier (SCC), which leverages visual-semantic contrastive learning coupled with a semantic regularization strategy to enhance class discrimination and semantic alignment.
Extensive experiments on three benchmark datasets (\textbf{SUN} \cite{patterson2012sun}, \textbf{AwA2} \cite{xian2018zero}, and \textbf{CUB} \cite{wah2011caltech}) demonstrate that our approach achieves performance comparable to state-of-the-art methods while using substantially fewer synthetic features.

The key contributions of this study include: (\textit{i}) We reformulate generative ZSL by drawing inspiration from FSL, shifting away from the conventional fully supervised paradigm,
(\textit{ii}) We propose a unified generative ZSL framework that synthesizes a limited set of group-level prototypes for unseen classes and addresses data imbalance using a semantic-aware contrastive classifier, and (\textit{iii}) We conduct comprehensive experiments and ablation studies on multiple benchmark datasets to validate the effectiveness and robustness of the proposed method.

\section{Related Works}
\label{sec:related_work}

\noindent \textbf{Embedding methods} retain the original training dataset and use samples from seen classes to learn a classifier that generalizes to unseen classes. These methods differ primarily in the space where classification is performed—visual, semantic, or a shared latent embedding space \cite{pourpanah2022review}. 
Visual embedding learns a projection from attributes to visual space, either by mapping semantic attributes into visual prototypes \cite{annadani2018preserving} or by leveraging semantic relationships between classes to build a classifier for unseen classes \cite{geng2021explainable, liu2020label}. Semantic embedding maps visual features into the semantic space \cite{zhang2019co}, while latent embedding maps both visual and semantic spaces into a shared latent space to bridge seen and unseen class gaps \cite{zhang2015zero}.

\noindent  \textbf{Generative methods} use seen class images and semantic information of both seen and unseen classes to synthesize visual features for unseen classes, which are then used to train classifiers for ZSL tasks. A key approach is GANs conditioned on unseen class semantics, such as f-CLSWGAN \cite{xian2018feature}, which generates visual features for unseen categories and matches them to semantic attributes in a joint embedding space. f-VAEGAN \cite{xian2019f} combines VAEs and GANs into a unified framework, while FREE \cite{chen2021free} enhances features within this framework to mitigate cross-dataset bias. ZeroGen \cite{ye2022zerogen} leverages pre-trained language models to synthesize datasets for zero-shot tasks. Recent works integrate embedding models with feature generation models to boost recognition across seen and unseen classes \cite{han2021contrastive, wu2024re, kong2022compactness}, such as CE-GZSL \cite{han2021contrastive}, which incorporates contrastive embedding with class- and instance-level supervision, and RE-GZSL \cite{wu2024re}, which extrapolates unseen class features using semantic relations with seen classes, contrastive losses, and a feature mixing module for more realistic and discriminative samples.

However, GAN-based generative methods often face challenges such as high computational costs \cite{guo2019deep} and mode collapse \cite{jahanian2019steerability}, and they do not inherently structure data representations in ways that are intuitive or easy to integrate with other models \cite{gowda2023synthetic}. Alternatively, several methods \cite{feng2022non, cavazza2023no, huynh2020compositional, lu2018attribute, guan2020zero, chou2020adaptive} synthesize samples without adversarial training. For example, methods like Composer \cite{huynh2020compositional} and ABS-Net \cite{lu2018attribute} extract and combine attribute-based features to create pseudo samples for unseen classes, while BPL \cite{guan2020zero} and AGZSL \cite{chou2020adaptive} use interpolation and perturbation to transfer class-specific variations from seen to unseen classes. Our approach aligns with these methods and synthesizes feature cluster centers of unseen classes as training data following a non-adversarial strategy.


 

 
\section{Approach}
\label{sec:approach}

\subsection{Problem Settings}
\label{secsec:problem_set}
In ZSL, there are \( K \) seen classes (\( \mathcal{Y}^s \)) and \( L \) unseen classes (\( \mathcal{Y}^u \)), with \( \mathcal{Y}^s \cap \mathcal{Y}^u = \emptyset \). Indices \( \{1, \ldots, K\} \) correspond to seen classes, and \( \{K+1, \ldots, K+L\} \) to unseen classes. The seen data comprises  \( N \) labeled images represented as \( D = \{(\mathbf{x}_i, y_i)|\mathbf{x}_i \in \mathcal{X}, y_i \in \mathcal{Y}^s\}_{i=1}^N \), while labeled samples for unseen classes are unavailable. The visual sample space is \( \mathcal{X} \), and class-wise semantic attributes \( \mathbf{A_o} = \{\mathbf{a}_c\}_{c=1}^{K+L} \) link seen and unseen classes. The goal in CZSL is to learn a classifier mapping \( \mathcal{X} \rightarrow \mathcal{Y}^u \), while GZSL aims for \( \mathcal{X} \rightarrow \mathcal{Y}^s \cup \mathcal{Y}^u \).

\begin{figure}[t]
  \centering
  \includegraphics[width=0.7\textwidth]{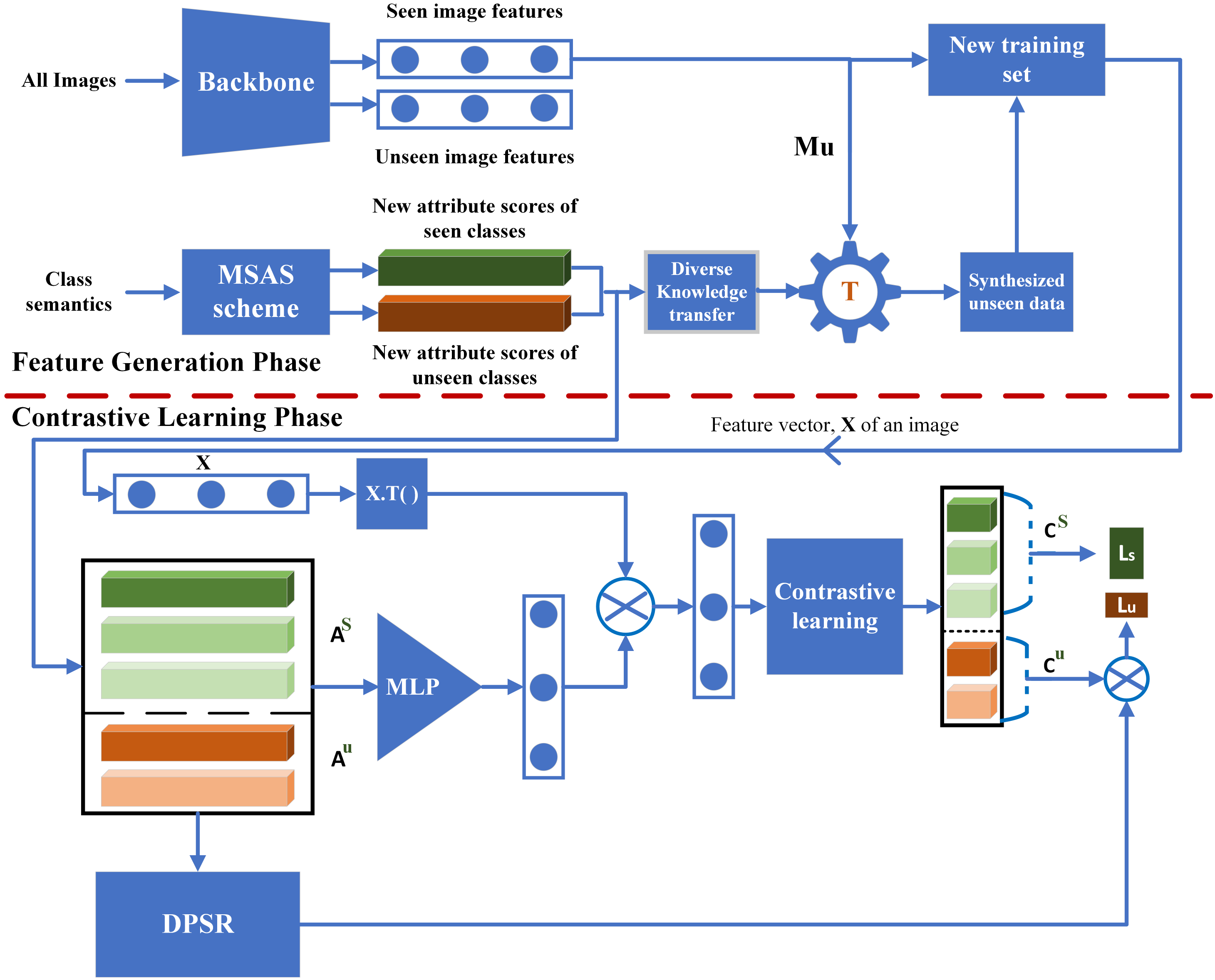}
  \caption{Overview of the FSIGenZ framework. The model comprises two main phases: feature generation (above the red dotted line) and contrastive learning (below). Symbols $\otimes$ and `T' represent element-wise multiplication and semantic knowledge transfer, respectively. Green-toned elements denote seen classes, while orange-toned elements denote unseen classes.}
  \label{fig:main_pipeline}
\end{figure}

\subsection{Model Overview}
\label{secsec:model_overview}
Fig. \ref{fig:main_pipeline} presents FSIGenZ, a unified zero-shot learning framework that seamlessly integrates feature generation and contrastive classification. In phase 1, it utilizes the MSAS scheme to re-score attributes and then estimates a compact yet semantically diverse set of subgroup prototypes per unseen class which capture intra-class variability without requiring unseen data. Instead of large-scale feature generation, these prototypes are included as synthetic training data for unseen classes. In phase 2, FSIGenZ employs the SCC unit built on visual-semantic contrastive learning along with a semantic regularization technique to address unseen class imbalance caused by limited feature generation. The complete process is detailed in Algorithm 1, with each component elaborated in subsequent sections.

\begin{figure}[t]
\captionsetup{type=algorithm, labelfont=bf,  textfont=normalfont, justification=raggedright, singlelinecheck=false}
\noindent\rule{0.77\textwidth}{0.6pt} 
\caption*{\textbf{Algorithm 1:} FSIGenZ}
\noindent\rule{0.77\textwidth}{0.6pt} 

\label{alg:algorithm}
\textbf{Input}: Images $I$, class attributes $\mathbf{A_o}$, initialization $\Theta^0$\\
\textbf{Output}: ZSL classifier
\begin{algorithmic}[1]

\STATE \textit{\# Extract Features}
\FOR{each image $i$}
    \STATE Extract ViT features: $F_i^{S+U} \leftarrow \text{ViT}(I_i)$
\ENDFOR
\STATE Apply MSAS on class attributes: $\mathbf{A} \leftarrow \mathbf{A_o}$

\STATE \# Synthesize Unseen Features
\STATE Compute relation matrices, $\alpha_p \leftarrow R_{s_p}$ for different $\lambda_p$ values
\STATE Estimate group clusters for unseen classes: $\mu^u_{k_p} \leftarrow M_s \alpha_p$
\STATE Utilize $\mu^u_{k_p}$ as synthetic unseen features
\STATE Obtain augmented train data, $F$ by combining synthetic unseen data and real train data

\STATE \textit{\# Train the Semantic-Aware Contrastive Classifier (SCC)}
\FOR{each iteration $t$}
    \STATE Encode class semantics with MLP: $E_j \leftarrow \text{MLP}(\mathbf{a_j})$
    \STATE Fuse features: $Z_{ij} \leftarrow F_i \otimes E_j$
    \STATE Compute class scores: $c_{ij} \leftarrow f(Z_{ij})$
    \STATE Apply Dual-Purpose Semantic Regularization (DPSR)
\ENDFOR

\STATE \textit{\# Perform Inference}
\FOR{a given test image $i$}
    \STATE Predict CZSL class: $y_i^{CZSL} \leftarrow \arg\max_j \{c_{ij}\}_{j=K+1}^{K+L}$
    \STATE Predict GZSL class: $y_i^{GZSL} \leftarrow \arg\max_j \{c_{ij}\}_{j=1}^{K+L}$
\ENDFOR

\end{algorithmic}
\noindent\rule{0.77\textwidth}{0.6pt}
\end{figure}

\subsection{Model-Specific Attribute Scoring (MSAS)}
\label{secsec:msas}

In general, attributes describe classes as a whole, and specifying them for individual instances is impractical. \textit{MSAS} balances between this impracticality and the need for instance-level specifications of attributes by tailoring them toward the model’s requirements. While one might treat each attribute as a hyperparameter, the brute-force approach of tuning every attribute as a hyperparameter is infeasible. Instead, MSAS redetermines attribute scores using collective measures such as thresholding and reweighting, which enable the model to utilize the most appropriate attribute values. In this paper, we employ a straightforward equation for MSAS:


\begin{equation}
  \mathbf{A} = (\mathbf{A_o} + \mathbf{A_{mdf}}) W_A, 
   \label{eq:1}
\end{equation}
where \( W_A \) is a weight scalar, \(\mathbf{A_o}\) is the original attribute matrix, and \(\mathbf{A_{mdf}}\) is defined as:

\begin{equation}
  \mathbf{A_{mdf}} = \mathbf{A_o} \odot (\mathbf{A_o} > T_h),
  \label{eq:2}
\end{equation}
where \(\odot\) denotes element-wise multiplication, and \((\mathbf{A_o} > T_h)\) is an indicator function producing a binary mask (1 if true, 0 otherwise). Adjusting hyperparameters \( W_A \) and \(T_h\) allows the model to align attribute scores with the optimal configuration.
\subsection{Unseen Data Synthesis}
\label{secsec:feature_synth}
Unlike conventional generative ZSL methods, our approach estimates subgroup prototypes of instances within each unseen class by modeling instance-level variations of attributes. These prototypes are then utilized as representative training samples. Inspired by MDP \cite{zhao2017zero}, we employ a multi-source knowledge transfer strategy to synthesize data for unseen classes by transferring the semantic embedding manifold into the visual feature space. To begin, we estimate virtual class centers for unseen classes using:
\begin{equation}
  A^u = R_s(A^s) \Rightarrow M^u = R_s(M^s),
   \label{eq:3}
\end{equation}
where \( A^u \) and \( A^s \) denote the attribute vectors for unseen and seen classes, respectively, \( M^u \) and \( M^s \) represent the mean cluster centers for unseen and seen classes, and \( R_s \) is the relation function mapping seen classes to unseen classes.

In this formulation, the attribute embeddings \( A^u \) of unseen classes are estimated by applying the relation function \( R_s \) to the seen class attributes \( A^s \). Subsequently, virtual class centers \( M^u \) for unseen classes are obtained by applying \( R_s \) to the cluster centers \( M^s \) of seen classes. The relation function \( R_s(\cdot) \) is learned through Sparse Coding, as:
\begin{equation}
  \min_{\alpha} \| a^u_c - A^s \alpha \|^2_2 + \lambda \| \alpha \|_2^2,
   \label{eq:4}
\end{equation}
where $\alpha = [\alpha_1, \ldots, \alpha_K]^T$ is the sparse coefficient vector, and $\lambda$ is the regularization parameter.

Finally, the estimated cluster center $\mu_{k}^u$ of the $k^{th}$ unseen classes is obtained as:
\begin{equation}
  \mu^u_k = M^s \alpha,
   \label{eq:5}
\end{equation}
where $\mu^u_k$ represents the estimated cluster center of the  $k^{th}$ unseen class.

While estimating a single cluster center per unseen class is relatively straightforward, the challenge arises when multiple centers must be calculated for each class. Addressing this complexity requires a solid theoretical foundation. Our observations, illustrated in Fig.~\ref{fig:msas}, offer crucial insights to support this approach.

Our method leverages dynamic attribute scores through two key mechanisms. The first involves our MSAS strategy (Sec.\ref{secsec:msas}), and the second builds on the idea of inducing variation by adjusting the regularization parameter $\lambda$ in equation \eqref{eq:4}, which in turn generates different sparse coefficient vectors $\alpha$. Substituting these diverse $\alpha$ values into equation~\eqref{eq:5} results in multiple cluster centers per unseen class. These centers act as visual prototypes, each capturing a subset of instances that share similar semantic traits, as shown in Fig.~\ref{fig:msas}(b). 

In our few-shot-inspired setup, this process of tuning $\lambda$ yields distinct advantages. It leads to varied, sparse representations that reflect meaningful subgroups within each class. By modeling dominant attribute combinations, the resulting prototypes represent intra-class diversity in a compact and informative manner. Importantly, this approach introduces minimal computational overhead, as only a limited number of prototypes are generated. In essence, the controlled variation of $\lambda$ offers an efficient and principled way to simulate the natural heterogeneity found within unseen classes—an idea that aligns with and supports the core design of FSIGenZ.



\subsection{Semantic-Aware Contrastive Classifier (SCC)} 
\label{secsec:SSC}
The Semantic-Aware Contrastive Classifier (SCC) begins by augmenting synthesized features of unseen classes with real features from seen classes, forming a unified training set. Let $F(\mathbf{x}_i)$ denote the feature representation of the $i^{th}$ instance, and $E(\mathbf{a}_j)$ represent the embedding of the semantic descriptor for class $j$. The fused feature $Z_{ij}$ is computed via a fusion operation, such as element-wise multiplication:
\begin{equation}
  Z_{ij} = F(\mathbf{x}_i) \otimes E(\mathbf{a}_j).
   \label{eq:6}
\end{equation}

To measure the compatibility between instance $i$ and class $j$, we define a contrastive score $c_{ij}$, computed as:
\begin{equation}
  c_{ij} = f(Z_{ij}),
   \label{eq:7}
\end{equation}
where $f$ is a contrastive learning function. 

Next, we formulate a one-hot class indicator vector $m_{ij}$, based on the ground-truth class label $y_i$ of instance $i$: 
\begin{equation}
m_{ij} = 
\begin{cases} 
1, & \text{if } y_i = j \\
0, & \text{if } y_i \neq j 
\end{cases}.
    \label{eq:8}
\end{equation}

Using these definitions, we can train a classifier to learn visual-semantic alignment by minimizing the following contrastive loss over the unified dataset, which contains both real and synthetic samples:
 \begin{equation}
      \mathcal{L_N} = - \sum_{i=1}^{N+N^S} \sum_{j=1}^{K+L} m_{ij} \log(c_{ij}) + (1 - m_{ij}) \log(1 - c_{ij}),
       \label{eq:9}
    \end{equation}
where $N^S$ is the number of synthesized features for unseen classes.

However, this formulation does not account for class imbalance, particularly among the unseen classes, where only a small number of synthetic instances are generated, consistent with our few-shot-inspired ZSL framework. To address this challenge, we introduce a semantic regularization strategy in the following section.



\textbf{Training Strategy:} To formulate the training objective, we partition the class indicator $m_{ij}$ and the contrastive score $c_{ij}$ into seen and unseen class components, denoted by superscript S and U, respectively. For each training sample, whether real (from seen classes) or synthetic (from unseen classes), we compute two separate losses. The first loss, $\mathcal{L_S}$, is a binary cross-entropy (BCE) loss applied to the model’s predictions over seen classes: 
    \begin{equation}
      \mathcal{L_S} = - \sum_{i=1}^{N+N^S} \sum_{j=1}^{K} m_{ij}^{s} \log(c_{ij}^{s}) + (1 - m_{ij}^{s}) \log(1 - c_{ij}^{s}),
       \label{eq:10}
    \end{equation}
    
Given that synthetic features are generated from a limited set of group prototypes, they may not fully reflect the complexity of real data. To mitigate this, we introduce a semantically regularized BCE loss $\mathcal{L_U}$, which operates on the model’s predictions over unseen classes. This loss incorporates a semantic similarity mask $s_{ij}^{u}$, which encodes the relationship between the ground-truth class and all other unseen classes. The formulation is: 
    \begin{equation}
      \mathcal{L_U} = - \sum_{i=1}^{N+N^S} \sum_{j=K+1}^{L} m_{ij}^{u} \log(c_{ij}^{u}s_{ij}^{u}) + (1 - m_{ij}^{u})\log(1 - c_{ij}^{u}s_{ij}^{u}),
       \label{eq:11}
    \end{equation}
    where $s_{ij}^u$ is a regularization term obtained by the Dual-Purpose Semantic Regularization (DPSR) module, discussed below. 
   The total training loss combines these components:
    \begin{equation}
      \mathcal{L} = \mathcal{L_S} + \beta \mathcal{L_U},
       \label{eq:12}
    \end{equation}
    where $\beta$ is a hyperparameter that controls the contribution of the regularized unseen-class loss.

\textbf{Dual-Purpose Semantic Regularization (DPSR):}
A key challenge in our ZSL formulation is the imbalance between seen data and synthetic unseen data. This disparity may lead the classifier to form overly confident decision boundaries for seen classes, while producing uncertain or unreliable predictions for unseen ones. To address this issue, we propose DPSR, a regularization strategy that enhances classifier training by incorporating semantic relationships between classes. The core idea is to guide the model using class-to-class semantic similarities. These similarities are obtained by solving the following optimization problem: 
\begin{equation}
        s_p = \arg\min_{s_p} \left\| \mathbf{a}_p - \sum_{q=1}^{K+L} \mathbf{a}_q s_{pq} \right\|_2^2 + \phi \| s_p \|_2,
       \label{eq:13}
    \end{equation}
 where \( \mathbf{a}_p \) represents the semantic embedding of class $p$, and $s_{pq}$ is the $q^\text{th}$ element of the similarity vector $s_p$, representing the semantic proximity between class $p$ and class $q$. The regularization parameter $\phi$ prevents trivial solutions by discouraging any single similarity score, particularly self-similarity, from dominating. Once the similarity vector is obtained, it is normalized to form a probability-like distribution:
    \begin{equation}
    s_{pq} = \frac{s_{pq}}{\sum_{q=1}^{K+L} \bar{s}_{pq}}.
    \label{eq:14}
    \end{equation}
This ensures that similarity scores across all classes are scaled appropriately

For any class $p$, the semantic similarity to any unseen class $q$ is denoted as $s_{pq}^u$, and is drawn from the computed vector $s_p$. These similarity values are used to modulate the model’s predictions for unseen classes within the contrastive loss function. In doing so, DPSR helps guide learning toward decision boundaries that are more consistent with the underlying semantic structure.
    

    \begin{figure}
    \centering
    \includegraphics[width=0.50\textwidth]{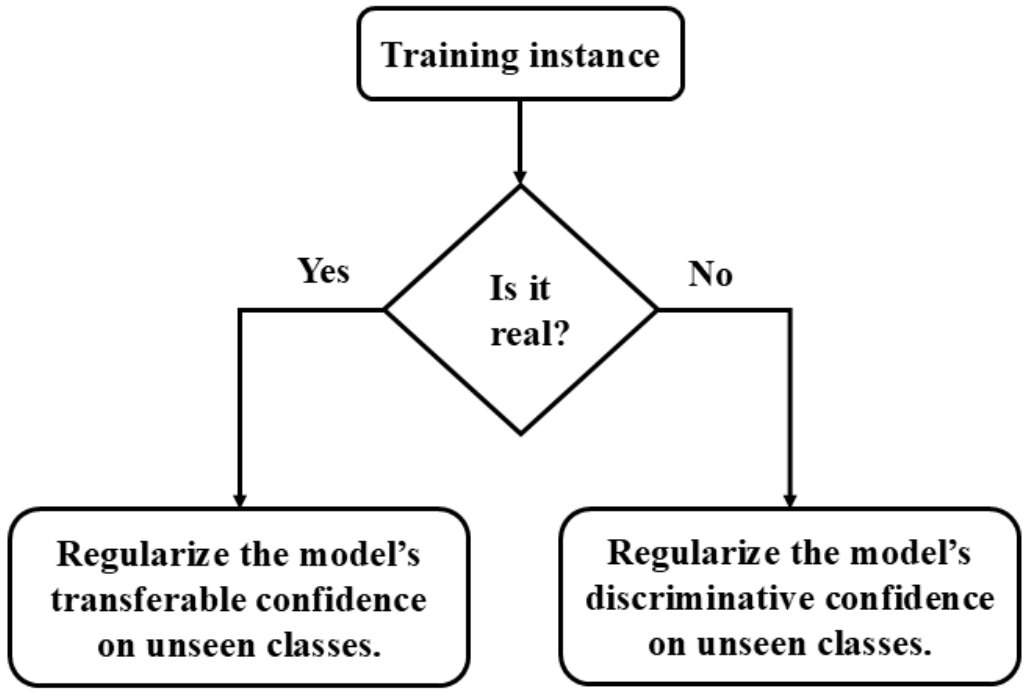}
      \caption {DPSR conceptualized in a flowchart: regularization of model's transferable and discriminative confidence to improve unseen class generalization.}
      \label{fig:DSPR}
    \end{figure}
    
DPSR tackles the challenge of data imbalance in unseen classes using a twofold strategy. First, for every training instance, the model generates prediction scores across both seen and unseen classes. For training instances belonging to seen classes, DPSR adjusts the transferable scores assigned to unseen classes by encouraging alignment with their semantic similarity to the true class. This helps promote reasonable activations for semantically related unseen classes while suppressing unrelated ones. Unlike existing approaches that adjust unseen scores post hoc \cite{huynh2020fine, xu2020attribute, zhu2019semantic}, this adjustment is integrated directly into the training process through semantic regularization. 
Second, for synthetic unseen instances, DPSR softens the discriminative loss signal among unseen classes to enhance robustness and generalization in the low-data regime. This regularization improves generalization by preventing the model from drawing overly sharp decision boundaries based on limited or imprecise data. Altogether, DPSR helps the classifier make more stable and semantically grounded predictions for unseen classes (see Fig.~\ref{fig:DSPR}). It is worth noting that DPSR is used exclusively during training; at inference, the model outputs raw predictions without any regularization.


\subsection{Zero-Shot Recognition}
We perform zero-shot recognition by computing contrastive scores between the visual representation of an input and the semantic embedding of all classes. In the CZSL, the prediction is restricted to unseen classes, and an input image is assigned to the class with the highest contrastive score among them: 
\begin{equation}
  \mathcal{P}_{czsl}(x_i) = \max_j \{ c_{ij} \}_{j=K+1}^{K+L}.
   \label{eq:15}
\end{equation}

In the GZSL setting, predictions are made over both seen and unseen classes. Accordingly, the image is assigned to the class (either seen or unseen) with the highest contrastive score:
\begin{equation}
  \mathcal{P}_{gzsl}(x_i) = \max_j \{c_{ij}\}_{j=1}^{K+L}.
   \label{eq:16}
\end{equation}

\section{Experimental studies}

This section presents the experimental setup, including the datasets, evaluation protocols, and implementation details. It then reports the experimental results and ablation studies.

\subsection{Experimental Setup}

\noindent \textit{\textbf{Datasets.}} We evaluate our approach on three ZSL datasets: SUN \cite{patterson2012sun}, AwA2 \cite{xian2018zero}, and CUB \cite{wah2011caltech}. AwA2 is a medium-scale, coarse-grained dataset with 37,322 images from 50 animal categories, described by 85 attributes. CUB is a fine-grained dataset with 11,788 images from 200 bird species, each with 312 attributes. SUN is another fine-grained dataset, comprising 14,340 images across 717 scene types with 102 attributes. 


\noindent \textit{\textbf{Baselines.}} We evaluate the performance of our proposed method against a wide range of existing approaches, including several generative-based models. Specifically, we compare with TCN \cite{jiang2019transferable}, DAZLE \cite{huynh2020fine}, ViT-ZSL \cite{alamri2021multi}, MSDN \cite{chen2022msdn}, SCILM \cite{ji2021semantic}, DUET \cite{chen2023duet}, BGSNet \cite{li2023diversity}, PRZSL \cite{yi2024prototype}, ZS-VAT \cite{10507018}, f-CLSWGAN \cite{xian2018feature}, f-VAEGAN-D2 \cite{xian2019f}, OCD-CVAE \cite{keshari2020generalized}, TF-VAEGAN \cite{narayan2020latent}, HSVA \cite{chen2021hsva}, TGMZ \cite{liu2021task}, GCM-CF \cite{yue2021counterfactual}, CE-GZSL \cite{han2021contrastive}, FREE \cite{chen2021free}, AGZSL \cite{chou2020adaptive}, SE-GZSL \cite{kim2022semantic}, ICCE \cite{kong2022compactness}, TDCSS \cite{feng2022non}, LCR-GAN \cite{ye2023learning}, DFCA-GZSL \cite{su2023dual}, RE-GZSL \cite{wu2024re}, AREES \cite{9881214}, JFGOPL \cite{li2024joint}, DENet \cite{ge2024towards} and DPCN \cite{10706204}. By benchmarking against this wide range of models, we ensure a comprehensive evaluation of our method's effectiveness in both ZSL settings.

\begin{table*}
\centering
\caption{Results of methods on CZSL (T1) and GZSL (H) tasks. Overall best and second-best results are bolded and underlined, while red and blue highlight the best and second-best among generative methods, respectively.}
\begin{adjustbox} {width=0.90\textwidth,center}
\begin{tabular}{ll@{\hskip 6pt}cccc@{\hskip 6pt}cccc@{\hskip 6pt}cccc}
\toprule
\multirow{2}{*}{} & \multirow{2}{*}{Method} & \multicolumn{4}{c}{SUN} & \multicolumn{4}{c}{AwA2} & \multicolumn{4}{c}{CUB}  \\
\cmidrule(lr){3-6} \cmidrule(lr){7-10} \cmidrule(lr){11-14} 
& & T1 & U & S & H & T1 & U & S & H & T1 & U & S & H  \\
\midrule
\multirow{9}{*}{\rotatebox{90}{\textbf{Embedding-based}}}
& TCN \cite{jiang2019transferable} &61.5 & 31.2 & 37.3 & 34.0 & 71.2 & 61.2 & 65.8 & 63.4 & 59.5 & 52.6 & 52.0 & 52.3  \\
& DAZLE \cite{huynh2020fine} & - & 52.3 & 24.3 & 33.2 & - & 60.3 & 75.7 & 67.1 & 65.9 & 56.7 & 59.6 & 58.1 \\
& ViT-ZSL \cite{alamri2021multi} & - & 44.5 & 55.3 & \underline{49.3} & - & 51.9 & 90.0 & 65.8 & - & 67.3 & 75.2 & \textbf{71.0} \\
& MSDN \cite{chen2022msdn}  & 65.8 & 52.2 & 34.2 & 41.3 & 70.1 & 62.0 & 74.5 & 67.7 & 76.1 &  68.7 & 67.5 & 68.1  \\
& SCILM \cite{ji2021semantic} & 62.4 & 24.8 & 32.6 & 28.2 & 71.2 & 48.9 & 77.8 & 60.1 & 52.3 & 24.5 & 54.9 & 33.8  \\
& DUET \cite{chen2023duet} & 64.4 & 45.7 & 45.8 & 45.8 & 69.9 & 63.7 & 84.7 & 72.7 & 72.3 & 62.9 &  72.8 & 67.5 \\
& BGSNet \cite{li2023diversity} & 63.9 & 45.2 & 34.3 & 39.0 & 69.1 & 61.0 & 81.8 & 69.9 & 73.3 & 60.9 & 73.6 & 66.7\\
& PRZSL \cite{yi2024prototype} & 64.2 & 53.6 & 37.7 & 44.4 & 73.6 & 65.8 & 77.8 & 71.3 & 77.1 & 68.8 & 63.7 & 66.2\\
& ZS-VAT \cite{10507018} & 62.6 & 45.6 & 33.8 & 38.8 & 72.2 & 59.9 & 80.8 & 68.8 & 75.2 & 67.5 & 68.1 & 67.8\\
\midrule
\multirow{21}{*}{\rotatebox{90}{\textbf{Generative-based}}} 
& f-CLSWGAN \cite{xian2018feature}   & 60.8 & 42.6 & 36.6 & 39.4  & -    & -    & -    & -    & 57.3 & 43.7 & 57.7 & 49.7  \\
& f-VAEGAN-D2 \cite{xian2019f} & 64.7 & 45.1 & 38.0 & 41.3  & 71.1 & 57.6 & 70.6 & 63.5 & 61.0 & 48.4 & 60.1 & 53.6\\
& OCD-CVAE \cite{keshari2020generalized}   & 63.5 & 44.8 & 42.9 & 43.8  & 71.3 & 59.5 & 73.4 & 65.7  & 60.3 & 44.8 & 59.9 & 51.3 \\
& TF-VAEGAN \cite{narayan2020latent}  & \underline{\textcolor{blue}{66.0}} & 45.6 & 40.7 & 43.0  & 72.2 & 59.8 & 75.1 & 66.6 & 64.9  & 52.8 & 64.7 & 58.1 \\
& HSVA \cite{chen2021hsva}     & 63.8 & 48.6 & 39.0 & 43.3  & -    & 56.7 & 79.8 & 66.3 & 62.8 & 52.7 & 58.3 & 55.3  \\
& TGMZ \cite{liu2021task}    & -    & -    & -    & -    & -    & 64.1 & 77.3 & 70.1 & -    & 60.3 & 56.8 & 58.5 \\
& GCM-CF \cite{yue2021counterfactual}   & - & 47.9 & 37.8 & 42.2    & -    & 60.4 & 75.1 & 67.0 & -    & 61.0 & 59.7 & 60.3 \\
& CE-GZSL \cite{han2021contrastive}   & 63.3 & 48.8 & 38.6 & 43.1 & 70.4 & 63.1 & 78.64 & 70.0 & 77.5 & 63.9 & 66.8 & 65.3 \\
& FREE \cite{chen2021free}     & - & 47.4 & 37.2 & 41.7  & - & 60.4 & 75.4 & 67.1 & - & 55.7 & 59.9 & 57.7 \\
& AGZSL \cite{chou2020adaptive}    & 63.3 & 29.9 & 40.2 & 34.3    & 73.8   & 65.1 & 78.9 & 71.3 & 57.2 & 41.4 & 49.7 & 45.2  \\
& SE-GZSL \cite{kim2022semantic}  & -    & 45.8 & 40.7 & 43.1 & -  & 59.9 & 80.7 & 68.8 & -    & 53.1 & 60.3 & 56.4 \\
& ICCE \cite{kong2022compactness} & - & - & - & - & 72.7 & 65.3 & 82.3 & 72.8 & 78.4 & 67.3 & 65.5 & 66.4  \\
& TDCSS \cite{feng2022non}    & - & - & - & - & - & 59.2 & 74.9 & 66.1 & -    & 44.2 & 62.8 & 51.9 \\
& LCR-GAN \cite{ye2023learning} & - & 57.6 & 43.8 & \textbf{\textcolor{red}{49.8}} & - & - & - & - & - & 53.6 & 67.5 & 59.7 \\
& DFCA-GZSL \cite{su2023dual}  & 62.6 & 48.9 & 38.8 & 43.3 & \underline{\textcolor{blue}{74.7}} & 66.5 & 81.5 & 73.3 & \underline{\textcolor{blue}{80.0}} & 70.9 & 63.1 & 66.8 \\
& RE-GZSL \cite{wu2024re} & - & - & - & - & 73.1 & 67.7 & 81.1 & \underline{\textcolor{blue}{73.8}} & 78.9 & 72.3 & 62.4 & 67.0 \\
& AREES \cite{9881214} & 64.3 & 51.3 & 35.9 & 42.2 & 73.6 & 57.9 & 77.0 & 66.1 & 65.7 & 53.6 & 56.9 & 55.2 \\
& JFGOPL \cite{li2024joint} & - & 48.8 & 38.0 & 42.7 & - & 62.6 & 74.2 & 67.9 & - &  56.4 & 62.7 & 59.4 \\
& DENet \cite{ge2024towards} & - & 52.3 & 40.8 & 45.8 & - & 62.6 & 84.8 & 72.0 & - & 65.0 & 71.9 & 68.3 \\
& DPCN \cite{10706204} & 63.8 & 48.1 & 39.4 & 43.3 & 70.6 & 65.4 & 78.6 & 71.4 & \textbf{\textcolor{red}{80.1}} & 72.7 & 65.7 & \textcolor{blue}{69.0} \\
& Zheng et al.\cite{zheng2025class} & - & - & - & - & - & 63.3 & 74.0 & 68.2 & - & 71.0 & 65.7 & 68.3 \\
& FSIGenZ (Ours) & \textbf{\textcolor{red}{67.8}} & 42.5 & 49.9 & \textcolor{blue}{45.9} & \textbf{\textcolor{red}{75.0}} & 67.6 & 82.3 & \textbf{\textcolor{red}{74.2}} & 73.0 & 65.9 & 72.7 & \underline{\textcolor{red}{69.1}} \\

\bottomrule
\end{tabular}
\end{adjustbox}
\label{tab:gzsl_comparison_main}
\end{table*}

\noindent \textit{\textbf{Evaluation Protocols.}}
We evaluate performance under CZSL and GZSL settings. In CZSL, we report the average per-class Top-1 accuracy (T1) on unseen classes. In GZSL, we compute Top-1 accuracies for seen (S) and unseen (U) classes, and use their harmonic mean $H = 2 \times \frac{S \times U}{S + U}$ to assess balanced performance across seen and unseen classes.

\noindent \textit{\textbf{Implementation Details.}} We extracted 786-dimensional image features using the ViT-Base \cite{dosovitskiy2020image} backbone pre-trained on ImageNet-1k and used attributes provided in \cite{xian2018zero}. In MSAS, we set \( W_A \) and \(T_h\) as \{0.005, 0.08, 0.3\} and \{0.7, 0.8, 0.7\} for SUN, AwA2 and CUB, respectively. For the required number of feature generation per unseen class of a dataset, $\lambda$ takes the same number of random values between 1 and 1.02. Our SCC unit is inspired by \cite{jiang2019transferable}, with a modified structure tailored to integrate both our underlying theoretical foundation and synthetic features from unseen classes. Specifically, our architecture includes a two-layer fully connected network to transform class semantics with 1024 and 786 units in the first and second layers, respectively. Contrastive learning is performed with a separate fully connected network featuring a 1024-dimensional hidden layer and a single-dimensional output. ReLU and Leaky ReLU are used for activation in the MLP, while ReLU and sigmoid functions are used in the contrastive network. The hyperparameter $\beta$ is adjusted empirically (see sensitivity analysis).

\begin{table*}
\centering
\caption{Unseen data statistics of different methods after including synthetic features. $N^{USF}$, $N^{URF}$, and $N^{SF}_P$ denote number of unseen synthetic features, number of unseen real features and number of synthetic features per class, respectively. (U/S) indicates whether the synthetic features are generated exclusively for unseen classes or for both seen and unseen classes.}
\begin{adjustbox} {width=\textwidth,center}
\begin{tabular}{lcccccccccccccccc}
\toprule
\multirow{2}{*}{Method} & \multicolumn{5}{c}{SUN} & \multicolumn{5}{c}{AwA2} & \multicolumn{5}{c}{CUB}\\
\cmidrule(lr){2-6} \cmidrule(lr){7-11} \cmidrule(lr){12-16}
 & T1 & H & $N^{USF}$ & $N^{URF}$ & $N^{SF}_P$ & T1 & H & $N^{USF}$ & $N^{URF}$ & $N^{SF}_P$ & T1 & H & $N^{USF}$ & $N^{URF}$ & $N^{SF}_P$ \\
\midrule
HSVA \cite{chen2021hsva} & 63.8 & 43.3 & 28800/14400 & 1440 & 400/200 (U/S) & - & 66.3 & 4000/2000 & 7913 & 400/200 (U/S) & 62.8 & 55.3 & 4000/2000 & 2967 & 400/200 (U/S)\\
CE-GZSL \cite{han2021contrastive} & 63.3 & 43.1 & 7200 & 1440 & 100 (U)  & 70.4 & 70.0 & 24000 & 7913 & 2400 (U) & 77.5 & 65.3 & 15000 & 2967 & 300 (U)\\
FREE \cite{chen2021free} & - & 41.7 & 21600 & 1440 & 300 (U)  & - & 67.1 & 46000 & 7913 & 4600 (U) & - & 57.7 & 35000 & 2967 & 700 (U)\\
ICCE \cite{kong2022compactness} & - & - & - & - & -  & 72.7 & 72.8 & 50000 & 7913 & 5000 (U) & 78.4 & 66.4 & 20000 & 2967 & 400 (U)\\
LCR-GAN \cite{ye2023learning} & - & 49.8 & 43200 & 1440 & 600 (U) & - & - & - & - & - & - & 59.7 & 20000 & 2967 & 400 (U)\\
RE-GZSL \cite{wu2024re} & - & - & - & - & -  & 73.1 & 73.8 & 50000 & 7913 & 5000 (U) & 78.9 & 67.0 & 20000 & 2967 & 400 (U)\\
DENet \cite{ge2024towards} & - & 45.8 & 7200 & 1440 & 100 (U)  & - & 72.0 & 35000 & 7913 & 3500 (U) & - & 68.3 & 10000 & 2967 & 200 (U)\\
DPCN \cite{10706204} & 63.8 & 43.3 & 5760 & 1440 & 80 (U)  & 70.6 & 71.4 & 25000 & 7913 & 2500 (U) & 80.1 & 69.0 & 15000 & 2967 & 300 (U)\\
Zheng et al.\cite{zheng2025class} & - & - & - & - & -  & - & 68.2 & 24000 & 7913 & 2400 (U) & - & 68.3 & 35000 & 2967 & 700 (U)\\
\textbf{FSIGenZ (Ours)} & 67.8 & 45.9 & 1080 & 1440 & 15 (U)& 75.0 & 74.2 & 900 & 7913 & 90 (U) & 73.0 & 69.1 & 500 & 2967 & 10 (U) \\
\bottomrule
\end{tabular}
\end{adjustbox}
\label{tab:gzsl_comparison_stat}
\end{table*}

\begin{table}
  \centering
    \caption{ The ablation study of the proposed method with (w.) and without (w/o) DPSR and MSAS. The best and second-best results are highlighted in red and blue, respectively.}
  \begin{adjustbox}{width=\columnwidth}
  \begin{tabular}{l c c c c c c c c c c c c}
    \toprule
    {Method} & \multicolumn{4}{c}{SUN} & \multicolumn{4}{c}{AwA2} & \multicolumn{4}{c}{CUB} \\
    \cmidrule(lr){2-5} \cmidrule(lr){6-9} \cmidrule(lr){10-13}
    & T1 & U & S & H & T1 & U & S & H & T1 & U & S & H \\
    \midrule
    
    Ours w/o MSAS w. DPSR & 64.6 & 36.3 & 46.3 & \textcolor{blue}{40.7} & \textcolor{red}{75.4} & 64.9 & 82.4 & \textcolor{blue}{72.6} & \textcolor{red}{73.4} & 63.3 & 71.2 & \textcolor{blue}{67.0}\\
    Ours w. MSAS w/o DPSR & \textcolor{blue}{65.3} & 12.0 & 46.0 & 19.1 & 69.7 & 8.8 & 93.3 & 16.1 & 46.9 & 16.2 & 59.5 & 25.5\\
    Ours w/o MSAS w/o DPSR & 64.5 & 12.5 & 45.1 & 19.6 & 72.0 & 11.0 & 95.8 & 19.8 & 62.8 & 20.6 & 54.0 & 29.9\\
    Ours w. MSAS w. DPSR & \textcolor{red}{67.8} & 42.5 & 49.9 & \textcolor{red}{45.9} &\textcolor{blue}{ 75.0} & 67.6 & 82.3 & \textcolor{red}{74.2} & \textcolor{blue}{73.0} & 65.9 & 72.7 & \textcolor{red}{69.1}\\
    \bottomrule
  \end{tabular}
  \end{adjustbox}
  \label{tab:ablation_study}
\end{table}



\subsection{Comparison With State-of-the-Art Methods}
Table \ref{tab:gzsl_comparison_main} presents a comprehensive comparison of recent methods on ZSL tasks for the SUN, AwA2, and CUB datasets. Specifically, FSIGenZ achieves the highest CZSL accuracy on SUN (67.8\%) and AwA2 (75.0\%), and also delivers competitive accuracy on CUB (73.0\%). In the GZSL setting, it achieves harmonic mean (H) scores of 74.2\% (AwA2), 69.1\% (CUB), and 45.9\% (SUN)—ranking as the best, second-best, and third-best method on these datasets, respectively. This level of consistent high performance across all datasets is unmatched by any other method in the table, many of which show strong results only in specific domains. Furthermore, methods that outperform FSIGenZ in GZSL on individual datasets—such as LCR-GAN (49.8\%) and ViT-ZSL (49.3\%) on SUN, and ViT-ZSL (71.0\%) on CUB—do not report CZSL performance, making it difficult to assess their overall effectiveness. Taken together, FSIGenZ stands out as the most consistently effective method in Table \ref{tab:gzsl_comparison_main}, considering ZSL performance across all benchmarks.

Compared to other generative-based methods, FSIGenZ achieves the best CZSL Top-1 accuracy on SUN (67.8\%) and AwA2 (75.0\%), as well as the best GZSL harmonic mean on AwA2 (74.2\%) and CUB (69.1\%), while securing the second-best harmonic mean on SUN (45.9\%), just behind LCR-GAN (49.8\%). However, despite LCR-GAN's higher harmonic mean on SUN, it falls short of FSIGenZ in CZSL Top-1 accuracy—indicating a weaker performance on the primary ZSL objective. A similar trend is observed on CUB, where DPCN records the highest CZSL accuracy (80.1\%), but FSIGenZ surpasses it in harmonic mean (69.1\% vs. 69.0\%). Moreover, FSIGenZ requires only 15, 90, and 10 synthetic features per class, amounting to 1080, 900, and 500 total synthetic features for SUN, AwA2, and CUB respectively—orders of magnitude fewer than other generative models, which often synthesize thousands of features (see Table \ref{tab:gzsl_comparison_stat}). For instance, LCR-GAN and DPCN synthesize up to 43200 and 15000 total unseen features on SUN and CUB respectively. Despite this, FSIGenZ outperforms or matches these methods in ZSL performance. This efficiency highlights a paradigm shift: instead of reframing ZSL as a fully supervised learning problem, FSIGenZ treats it more like a FSL problem by generating only a handful of informative synthetic features that capture the essential structure of unseen classes. 

\subsection{Ablation and Sensitivity Analysis} 
\noindent \textit{\textbf{Ablation Study.}} In this section, we evaluate the impact of different components of the proposed method. We systematically remove various components—namely, MSAS and DPSR unit—from our method (see Table \ref{tab:ablation_study}). Our full model (last row), which integrates both MSAS and DPSR, achieves the best overall performance across all datasets. With DPSR removed, we observe a significant drop in GZSL performance, which indicates that the DPSR is crucial for generalization. Removing MSAS while retaining the DPSR leads to a more balanced performance across both CZSL and GZSL tasks, highlighting the DPSR's role in maintaining prediction consistency. However, when both components are removed, the model's performance declines significantly across all datasets, which confirms that both MSAS and the DPSR are essential for achieving high accuracy and generalization.

\begin{figure}[!htbp]
  \centering
  \includegraphics[width=0.95\textwidth]
  {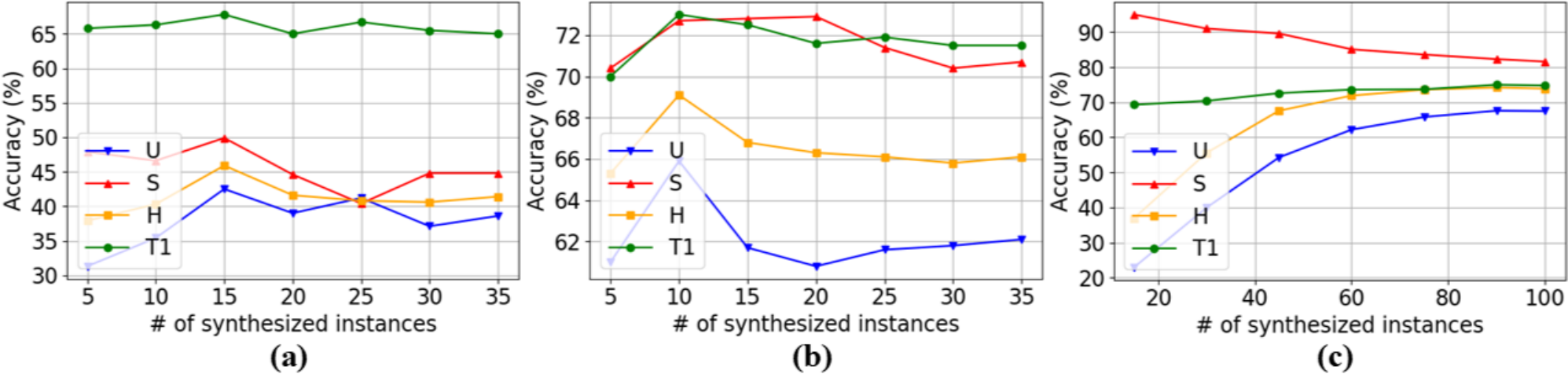}
  \caption{Results concerning varying synthetic instances for each unseen class of the (a) SUN, (b) CUB, and (c) AwA2 datasets.}
  \label{fig:graph_all_datasets}
\end{figure}

\begin{figure}[!htbp]
  \centering
  \includegraphics[width=0.95\textwidth]
  {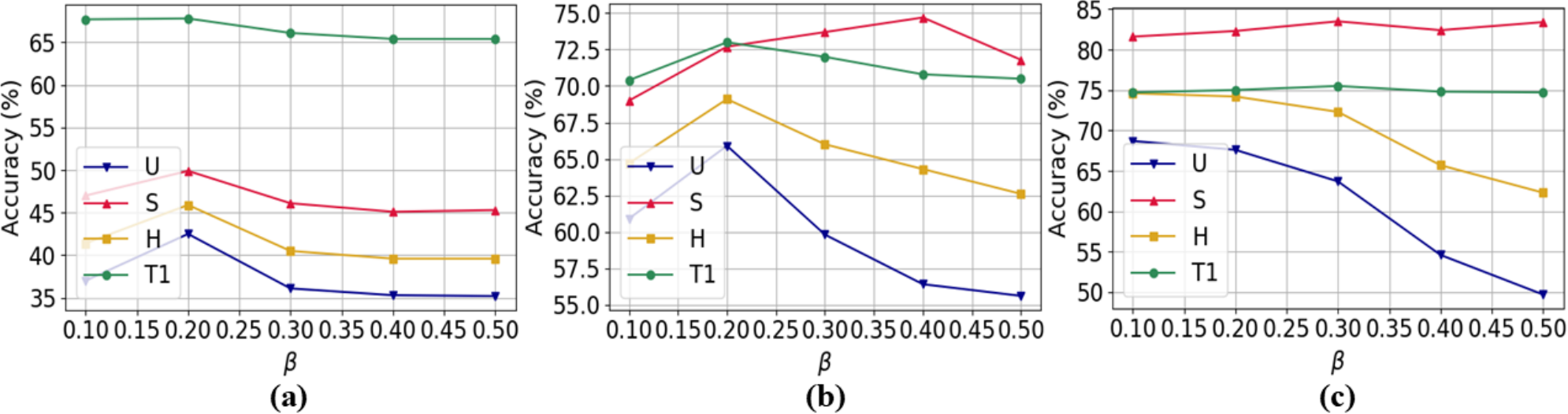}
  \caption{Results for various $\beta$ values on (a) SUN, (b) CUB, and (c) AwA2 datasets.}
  \label{fig:beta_all_datasets}
\end{figure}

\noindent \textit{\textbf{Impact of synthesized instances and  $\beta$.}} We test the impact of varying the number of synthesized instances per unseen class to understand how this factor influences the model's ability to generalize across different datasets. As shown in Fig. \ref{fig:graph_all_datasets} (a-c), increasing the number of these instances affects the model's performance and generalization capabilities and provides valuable insights into the optimal balance needed to achieve robust performance.
In addition, we analyze the effect of $\beta$ on model performance for each dataset (see Fig.~\ref{fig:beta_all_datasets} (a–c)). The results indicate that our method achieves optimal performance on all the datasets with $\beta$ set to 0.2.

\vspace{-10pt}
\begin{figure}[!htbp]
  \centering
  \includegraphics[width=0.95\textwidth]
  {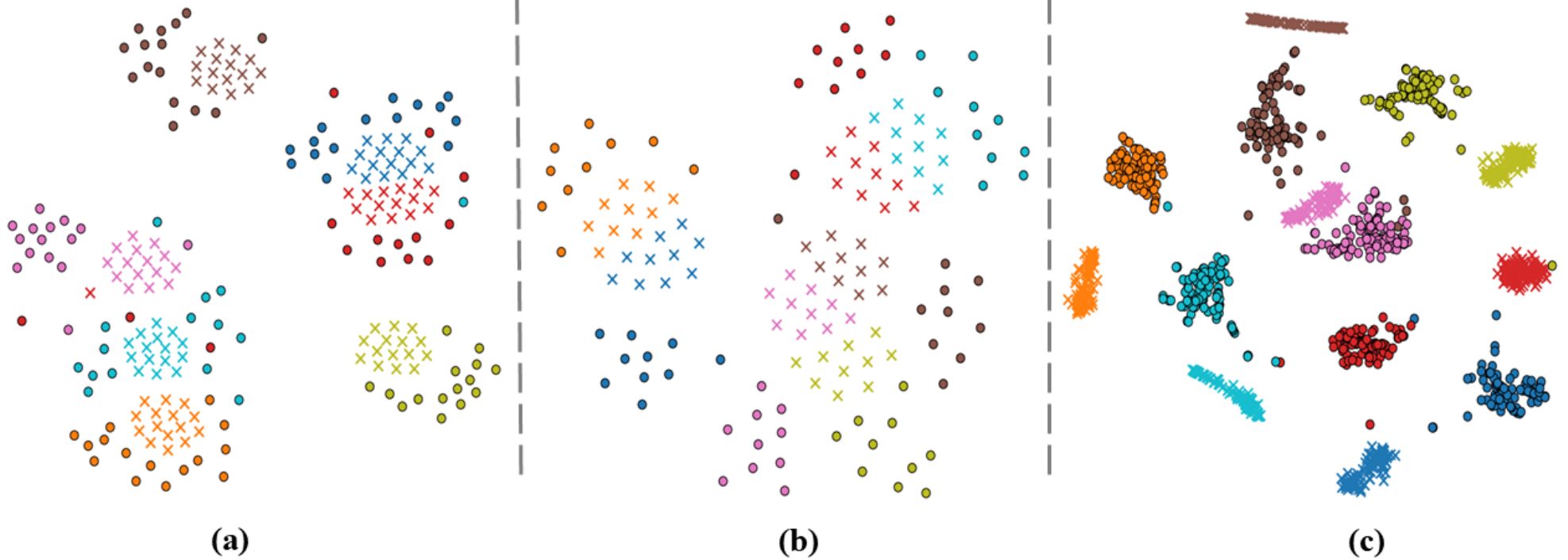}
  \caption{t-SNE visualization of seven randomly selected real (circle) and estimated (cross) clusters of the (a) SUN, (b) CUB and (c) AwA2 datasets.}
  \label{fig:tsne_all_datasets}
\end{figure}

\noindent \textit{\textbf{Visualizing Subgroups in Unseen Class Space.}}
To assess how well the synthesized prototypes represent the internal structure of unseen classes, we visualize real and estimated subgroup centroids using t-SNE on all three datasets (Fig.~\ref{fig:tsne_all_datasets}). Real sub-clusters (colored circles) are obtained via k-means on image features, while estimated prototypes (colored crosses) are generated by FSIGenZ. On SUN, despite its high intra-class variability, estimated prototypes align well with real clusters. In CUB, alignment remains strong despite its fine-grained nature. On AwA2, the match is especially close, reflecting its coarse-grained structure and strong attribute signals. These results highlight FSIGenZ’s ability to infer meaningful semantic subgroups using only class-level attributes.
\section{Conclusion}
    \label{sec:conclusion}

In this paper, we present FSIGenZ, a unified zero-shot learning framework that combines feature generation and contrastive classification into a cohesive system. FSIGenZ synthesizes a compact set of semantically diverse subgroup prototypes per unseen class to avoid large-scale data generation. By modeling instance-level attribute variability, it estimates these prototypes and uses them as training data. To address class imbalance, a semantic regularization strategy is incorporated during the training of a contrastive classifier. FSIGenZ achieves competitive performance on SUN, CUB, and AwA2 datasets with significantly fewer synthetic features. By adopting a few-shot-inspired perspective, FSIGenZ reduces computational overhead and better preserves the original spirit of ZSL. In our future work, we will explore adaptive subgroup modeling to enhance performance in fine-grained domains.

\bibliographystyle{unsrt}  
\bibliography{references}  






\end{document}